\documentclass[letterpaper, 10 pt, conference]{ieeeconf}  % Comment this line out if \usepackage[nocompress]{cite}
\usepackage[pdftex]{graphicx}
\usepackage{dblfloatfix}
\usepackage{amsmath}
\usepackage{algorithmic}
\usepackage{array}
\usepackage{float}
\usepackage[caption=false,font=footnotesize,labelfont=sf,textfont=sf]{subfig}
\usepackage{url}
\usepackage{amsfonts}
\usepackage{amssymb}
\usepackage{adjustbox}
\usepackage{multicol}
\usepackage{tabularx}
\usepackage{makecell}
\usepackage{verbatim}
\usepackage{caption}
\usepackage{hyperref}
\usepackage{xspace}
\usepackage[dvipsnames,svgnames,x11names]{xcolor}
 \usepackage[subtle,tracking=normal]{savetrees}

\newcommand{\tightsection}[1]{\vspace{-0.05in}\section{#1}\vspace{-0.02in}}

\newcommand{\mypara}[1]{\smallskip\noindent{\bf {#1}:}~}
\IEEEoverridecommandlockouts                              % This command is only needed if 
\def\etal{\mbox{\textit{et al.}}\@\xspace}

\title{\centering \LARGE \bf Enhancing Haptic Distinguishability of Surface Materials\\ with Boosting Technique}
\author{Priyadarshini~K$^{1,2}$ and Subhasis Chaudhuri$^{1}$ 
\thanks{$^{1,2}$Priyadarshini K is with Department of Electrical Engineering, IIT Bombay and Sony AI, Tokyo {\tt\small priyadarshini.kumari@sony.com}}
\thanks{$^{1}$Subhasis Chaudhuri is with Department of Electrical Engineering, IIT Bombay {\tt\small sc@ee.iitb.ac.in}}      
\thanks{We would like to thank Siddhartha Chaudhuri for his useful feedback that helped in improving the paper. } }

\begin{document}

\maketitle
\thispagestyle{empty}
\pagestyle{empty}

\begin{abstract}

Discriminative features are crucial for several learning applications, such as object detection and classification. Neural networks are extensively used for extracting discriminative features of images and speech signals. However, the lack of large datasets in the haptics domain often limits the applicability of such techniques. This paper presents a general framework for the analysis of the discriminative properties of haptic signals. We demonstrate the effectiveness of spectral features and a boosted embedding technique in enhancing the distinguishability of haptic signals. Experiments indicate our framework needs less training data, generalizes well for different predictors, and outperforms the related state-of-the-art.

\end{abstract}

\tightsection{Introduction} \label{Intro}

Different sensory inputs (vision, smell, touch, etc.) are crucial to endow robots with human-like understanding of the physical world. For instance, when we hold an object, we see its color and shape, feel its texture, and smell its scent; our brain binds these perceptions together to form the concept of that object. While there have been extensive studies in the vision domain, the learning problem is relatively unexplored in the haptics domain.

For many robotics applications, the haptic sensation is crucial for manipulating objects in the surrounding. As a result, a great body of research in recent years~\cite{TUM,Katherine14,gao2016deep,strese2019haptic,Sinapov} have focused on understanding tactile properties such as surface friction, compliance (soft vs. hard), and the fragility of objects. Besides haptic adjectives, surface material classification is also relevant for robot identifying, grasping, and characterizing objects. Existing methods of classifying tactile properties of objects either use input signals recorded from very high-end sensors~\cite{Chathuranga} or rely on multimodal features for surface material classification~\cite{TUM,Katherine14,gao2016deep,strese2019haptic}. While Sinapov \etal~\cite{Sinapov} use single modality acceleration data to generate discriminative features for material classification, they achieve very low classification accuracy (\mbox{$\sim$30-40\%}). The accuracy improves only with the inclusion of other modalities such as image, sound, friction, and scanning velocity~\cite{TUM,gao2016deep}. However, this requires additional sensing hardware for data collection, and that reduces the generalizability of algorithms. Moreover, the feature vectors in some of these works are also quite high-dimensional (e.g., 125-D in ~\cite{Sinapov}), which can lead to overfitting. We present a metric-based feature transformation technique that learns a set of discriminative tactile features by projecting {\em single} modality acceleration data of surface materials to an embedding space where the distinguishability between different classes of haptic signals is well realized.

Most often, in the raw input space, no underlying pattern is easily visible within the data. The data is typically too high-dimensional, and different classes of signals are not well-separated in this space. Neural network models are commonly used to capture the complex underlying relationship in the data. However, the success of such methods can be largely attributed to the powerful computing systems and abundance of training data. The latter is often a limiting factor in the haptics domain. Moreover, highly parameterized techniques such as neural networks lack generality, adaptability, and robustness
in the case of smaller datasets.

We propose a {\bf general computational framework}, founded on {\bf spectral features} for haptic textures generated only from acceleration data, augmented by a {\bf boosted embedding} technique trained with discrete label-based supervision. Such embedding has several important benefits. The performance of various pattern recognition algorithms, such as classification and clustering, can improve significantly after re-scaling the data using a learned embedding ~\cite{andrew}. Depending upon how the embedding function is learned, the projected data in the learned space reflects the distinguishability between signals. Our learning framework integrates two crucial components: (1) discriminative feature extractor, (2) metric-based feature transformer. We apply the spectral transformation to the input signals and extract representative components using the binning operation to learn discriminative compact features from high-dimensional acceleration signals. We call this constant Q-factor filter bank (CQFB) feature, and it's briefly described in \cite{priyadarshini2019perceptnet}. Next, we improve the distinguishability of haptic signals by transforming the extracted features using a linear projector called BoostMetric\cite{boost}. Our experiments demonstrate that the CQFB feature augmented with the boosted embedding gives much better performance of semantic discrimination tasks such as classification and clustering than without embedding. 
\tightsection{Related Work}\label{RW}

Distinguishing between surface materials using haptic touch is difficult due to the presence of high-frequency vibrations component in the texture signals. Several prior works use different discriminative textural features such as roughness, friction, and hardness of materials for classification tasks \cite{gao2016deep,strese2019haptic,TUM,WHC15,Culbertson,Sinapov,richardson2019improving}. Strese \etal~\cite{strese2019haptic,TUM,gao2016deep} use multi-modal features (tactile and visual) to improve the performance of the classifier. Authors in \cite{Zheng,gao2016deep} employ the powerful capability of the deep neural network to learn automated discriminative features, again using multi-modal sensory inputs. Joseph \etal~\cite{Katherine14} and Strese \etal~\cite{TUM} report very low classification accuracy  (\mbox{$\sim$30-40\%}) when haptic data (acceleration signal) alone is used for feature extraction. Joseph \etal~\cite{Katherine14} and Jeremy \etal~\cite{Fishel} show that successful recognition of surface material depends on scanning velocity and normal force applied during recording. Strese \etal~\cite{WHC15} mitigates this effect by proposing a deep neural network approach that learns features from acceleration and tapping data. Although the input is just a haptic modality, acceleration trace and tapping data require a separate recording procedure.

Another exciting line of work \cite{Sinapov,Fishel,lederman90haptic} shows along with tactile information, the prior knowledge of exploratory movements is also needed to characterize the objects. A study by Lederman \etal~\cite{lederman90haptic} shows significant improvement in KNN classification accuracy (from 48$\%$ to 68$\%$) using multiple exploratory movements. Further, Jeremy \etal~\cite{Fishel} build an advanced model that adaptively selects the optimal feature and the optimal exploratory movements. Moreover, they use a high-end BioTac sensor for haptic data recording, which has similar mechanical properties as a human fingertip. They report a classification accuracy of around 95$\%$ on 117 textures.
In contrast to our approach, all methods use either multi-modal or high-end data recording sensors such as a biomimetic sensor and multiple exploratory movements for feature generation. Our approach tries to classify 69 surface materials using the {\bf acceleration data alone}, which is recorded using a low-cost acceleration sensor (LIS344ALH) in a {\bf single exploration}. The dimension of our feature vector is also low (only 10-13). Moreover, our method provides a formal and general representation of haptic textures that directly captures the relationship between different signals in terms of similarity or dissimilarity.

The closest work to ours is \cite{phonemes}, where embedding space is learned using MDS-based projection approaches. While these technique has been widely used for several applications including clustering and modeling perceptual dissimilarity between objects, they are non-parametric and cannot compute a suitable projection of a new signal with new features at test time.
\tightsection{Method}\label{Method}

In this section, we present the various stages of our framework. Our objective is to learn the metric-based feature transformation function over the input space to enhance the distinguishability of haptic textures. Our data consists of traces of acceleration signals of different surface materials such as stone, wood, grass, paper, etc.  The learned transformation function re-scales the data in the embedding space such that the dissimilar haptic signals (say stone and grass) are placed far apart, and a similar pair (say stone and granite) are placed close together.

To reduce the dimension of data, we combine the three components of the acceleration signal $y(t)$ into a single vector representing the spectral magnitude $Y(w)$ using DFT321 \cite{DFT321}. As mentioned in \cite{DFT321}, the method preserves the spectral characteristic of the acceleration signal even after reducing the dimension. The psychophysical studies show that touch is highly sensitive up to 1 kHz frequency, and the optimal sensing frequency is around 250 Hz \cite{optimal_perception_freq}.
We thus have taken a signal spectrum $Y(w)$ within the frequency of 1 kHz for feature extraction. Each component of our algorithm is explained in the following sections.

\mypara{Feature extraction}\label{Feature_extraction}
The major problem in learning discriminative features with time series data is the phase invariant compact representation. Moreover, the distance metric used for feature transformation heavily relies on the meaningful feature to place similar data close to each other and dissimilar data far apart. However, in a high dimensional space, like in the case of time-series data, the contrast between similar and dissimilar data decreases \cite{freqFeature}. Therefore it is essential to select the right feature set which can compactly represent the input data, keeping only important information while discarding the noise and redundant data. The frequency-domain analysis of texture signals is useful in terms of the reduction of correlated data or noise. Hence in our learning algorithm, we use spectral features.

\begin{center}%[thpb]
\captionsetup{type=figure}
\includegraphics[scale=0.3]{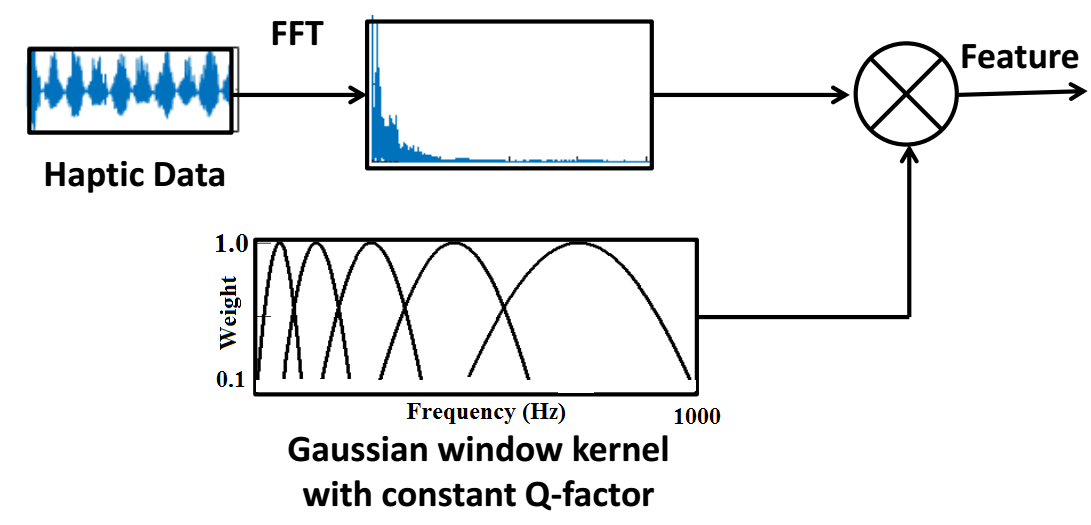}
\caption{A variable bandwidth Gaussian shaped filter bank with constant Q-factor is used for feature extraction.}
\label{feature}
\end{center}
We extract the features from the signal spectrum $Y(w)$ by dividing them into  $N$ bins of varying sizes. Each component of the feature vector indicates the total spectral energy within each bin. Since the human sensitivity to vibration falls exponentially as frequency increases \cite{psychophysics1}, we propose to vary the bin size in geometric progression as one moves from low frequency to high frequency (as shown in Fig. \ref{feature}). Thus, low-frequency components get more prominence than high-frequency components and reduce the effect of noise in the data. Further, instead of using a rectangular window to integrate the spectral components within a bin, we use a  window with an appropriate width of $\sigma$ to provide a certain spectral overlap between adjacent bins.

We apply a variable bandwidth Gaussian shaped filter bank $\{W_{j}(w)\}_{j=1}^{N}$ on frequency spectrum of the signal $Y(w)$ as shown in Fig.~\ref{feature}. The sigma($\sigma$) and center frequency ($f_{c}$) of Gaussian kernel is varied in geometric progression with a constant \mbox{Q} factor $\alpha$.  At $\alpha$ equal to one, we get a regularly spaced, constant bandwidth spectral bin.
In order to keep the Q-factor constant, we vary the center frequency ($f_{c}$) as well as $\sigma$ of the Gaussian kernel in geometric progression as follows-
\begin{equation} \label{variableFC1}
f_{c_{j}} = \alpha^{j-1}f_{c_{1}}; \hspace{0.5cm}\sigma_{j} =\alpha^{j-1}\sigma_{1}; \hspace{0.3cm} j \in[1, N]
\end{equation}
where $N$ denotes the dimension of the feature vector. Since the central frequency of the last Gaussian window($f_{c_{N}}$) cannot exceed the maximum frequency of spectrum ($f_{max}$), we have fixed  $f_{c_{N}}$  equal to  $f_{max}$. Thereafter we compute other required parameters of the Gaussian filter bank as follow-
\begin{enumerate}
\item First, we find the center frequency of the first Gaussian window ($f_{c_{1}}$) using Eq. \ref{variableFC1}, \textit{ie,} $f_{max} = \alpha^{N-1}f_{c_{1}}$.
\item Now we compute the sigma of the first Gaussian window ($\sigma_{1}$) such that its passband dies down to $10\%$ level at zero frequency, \textit{ie,}
$e^ - {\frac{f_{c_{1}}^{2}}{2\sigma_{1}^2}} = 0.1$.
\end{enumerate}
Having estimated $f_{c_{1}}$ and $\sigma_{1}$, one can find $f_{c_{j}}$ and $\sigma_{j} \quad \forall \quad j$ using Eq.\ref{variableFC1}. One may note that $\frac{f_{c_{j}}}{\sigma_{j}}$ is the same for all $j$ and hence the $Q$ factor is constant. The computed $j^{th}$ component of feature is given by
\begin{equation} \label{feature_Comp}
a_{j} = \int_{f_{j-1}}^{f_{j}} W_{j}(w)|Y(w)|^{2} dw  \hspace{0.3cm} j \in[1, N]
\end{equation}
We call the derived feature $x = \{a_{1}, a_{2},.....a_{N}\}$ as constant Q-factor bandwidth (CQFB) feature. The next step of our algorithm is to learn an appropriate metric-based feature transformation for better distinguishability of haptic textures.

\mypara{Feature transformation using boosting technique} \label{embedding}
Our goal is to learn a feature transformation matrix $M \in \mathbb{R^{N \times N}}$ such that in the projected space, the Euclidean distance between two haptic signals indicates the dissimilarity between them. We use the boosting-like technique proposed by Chunhua \etal~\cite{boost} for linear metric-based feature transformation. The idea is based on the interesting insight that any positive semi-definite matrix such as Mahalanobis distance matrix can be decomposed into a linear positive combination of trace-one-rank-one matrices. Therefore an efficient metric can be learned efficiently by combining multiple weak learners well-suited for small datasets.

Estimating the Mahalanobis distance $d_{M}(x_{i}, x_{j})$ is equivalent to first transforming the data into a new space using a linear projector $x \rightarrow L^{T}x$ and then applying the Euclidean metric to the transformed data. One of the major challenges in learning the Mahalanobis matrix is to ensure the positive semi-definiteness constraint on the learned matrix $M$. To ensure positive semi-definiteness, sometimes transformation matrix $\mathbf{L}$ is learned instead of $\mathbf{M} = LL^{T} \in \mathbb{R}^{N \times N}$ \cite{NCA, LMCA}. But this approach does not ensure the global optimum value as the cost function becomes non-convex in this case. The standard SDP (semidefinite programming) solver can also be used to ensure the global optimum, but SDP is inefficient for metric learning, especially when the feature dimension is high. The boosting technique~\cite{boost} addresses this challenge by incorporating the properties of a positive semi-definite matrix in its learning approach itself. The method combines the trace-one and rank-one matrix iteratively, and the final metric $\mathbf{M} \in \mathbb{R}^{N \times N}$ takes the form of $ \mathbf{M} = \sum_{k = 1}^{K} w_{k}\mathbf{U_{k}}$. The $w_{k}$ is the non-negative  weight and the matrix $\mathbf{U_{k}}$ is a trace-one, rank-one matrix. In this way, the method incorporates the positive-definiteness constraint elegantly without increasing the computational complexity.

The input to the boosting algorithm requires a set $\textit{C}$ of order relations among data.
\begin{equation}\label{const set}
\mathbf{C} = \{(x_{i}, x_{j}, x_{k})  \hspace{0.2cm}| \hspace{0.2cm} d_M (x_{i}, x_{j}) < d_M (x_{i}, x_{k})\}
\end{equation}

Here the training samples  $x_{i}$ and $x_{j}$ belong to the same class, and $x_{k}$ belongs to a different class. The set of constraints $\mathbf{C}$ is in the form of triplets of training samples where the sample $x_{i}$ is closer to sample $x_{j}$ than $x_{k}$. In order to create well separated compact clusters of different classes, we maximize the distance margin $\rho_{r}=(x_{i} - x_{k})^{T}\mathbf{M}(x_{i} - x_{k}) -  (x_{i} - x_{j})^{T}\mathbf{M}(x_{i} - x_{j})$ between the pair $(i, j)$ and $(i, k)$ of each triplet $r$ in the Mahalanobis distance metric space.

As mentioned earlier, the aim is to maximize the distance margin ($\rho_{r}$) with an appropriate  regularization. The optimization problem is defined~\cite{boost} using an exponential loss function as follow
\begin{equation} \label{primal}
\begin{aligned}
&\underset{\mathbf{M}}{\text{minimize}}
\hspace{0.5cm} \log(\sum_{r =1}^{|\mathbf{C}|}\exp(- \rho_{r})) + v\mathbf{Tr(M)}\\
& \text{subject to} \hspace{0.5cm} \mathbf{M} \succeq 0.
\end{aligned}
\end{equation}
Here $v$ is a regularizing parameter. The regularizer term prevents the optimization problem leading to learning an arbitrarily large $M$ to make the exponential loss zero. Note the in the boosting technique distance metric $M$ is learned by incrementally adding the rank-one and trace-one matrices $U_{k}$ as bases learners. The crucial part of solving the optimization problem \ref{primal} is estimating all basis learners $U_{k}$ and the number of such possible $U's$ is infinite. Therefore, instead of directly solving the primal problem \ref{primal}, the column generation technique \cite{colGen} is used, which finds the approximate solution by estimating the subset of entire variables set at each step.  The new variable is learned incrementally only if it improves the solution significantly. The details of the algorithm can be found in \cite{boost}.

\tightsection{Experiments}\label{sec:experiments}

We conduct various experiments to evaluate the accuracy and effectiveness of our algorithm. Firstly, we demonstrate how the defined CQFB feature and the learned embedding improve the performance of texture classification tasks. In each case, we compare the performance to those alternatives from ablation studies or prior work. To ensure the comparability of different methods, the same test data is used for all tested methods. Our implementations are in Matlab, and we used the reference code from authors' websites as appropriate. Our data are drawn from the TUM texture repository~\cite{TUM}. The dataset has acceleration data corresponding to 69 different surface materials. The recording of training and testing data is done separately by different users. The training data is recorded by domain experts, whereas testing data is recorded by subjects unfamiliar with work details. Each training and testing data contains 690 samples, ten samples in each class. In the prior work \cite{TUM}, the algorithm is evaluated using 69 different surface materials. The detail of the data recording procedure is described in~\cite{TUM}. We treat the samples within a class as equivalent, and randomly pick a representative every time we need features for a class. For all experiments below, testing is done on a separately recorded testing dataset, which is different from the training dataset.

\begin{table}[ht]
\centering
\begin{tabular}{|c c c c|}
 \hline
 Feature & Accuracy  & Accuracy & Ref. \\
  & w/o boosting (\%) & with boosting (\%) & \\
 \hline
 Spectral Peak & 12 & 12 & \cite{Jamali}\\
 Spectral frequency  & 2 &  4 & \cite{Jamali}\\
 Energy Values  & 43 & 65 & \cite{Katherine14}\\
CQFB($\alpha = 1$) & 2.5 & 13 & Ours\\
CQFB($\alpha = 1.8$) & $\mathbf{60}$ & $\mathbf{82}$ & Ours\\
\hline
\end{tabular}
\caption{Comparison of $k$-NN classification performance with different features. The results show the improvement in classification accuracy for different features when the boosted embedding is used in conjunction.}
\label{accuracyTable1}
\end{table}

\begin{table}[ht]
\centering
\begin{tabular}{|c c c c c c|}
 \hline
 Feature & Data  & Dim & Bayes & Decision  & Ref. \\
   &   &  &  &  Tree &  \\
\hline
LPC &  Acc & 10 & 5 & 4 & \cite{Culbertson}\\
Spectral  &  Acc & 5 &  9 & 9 & \cite{Jamali}\\
peaks &   &  &   &  & \\
Roughness &  Acc & 1 & 11 & 10 & \cite{Fishel}\\
Energy & Acc & 30 &  29 & 30 & \cite{Katherine14}\\
Spectrogram & Acc & 125 & 39 & 29 & \cite{Sinapov}\\
Multimodal & Acc, Img & 6 & $\mathbf{74}$ & 67 & \cite{TUM}\\
features &  Snd, Frc &  &  &  & \\
CQFB($\alpha = 1.8$) & Acc &  11 & $72$ & $\mathbf{69}$ & Ours\\

\hline
\end{tabular}
\caption{Comparison of classification accuracy ($\%$) of our approach and related work by Strese~\etal \cite{TUM}. The results of related work are taken  from the corresponding paper \cite{TUM}.\protect\footnotemark }
\label{accuracyTable2}
\end{table}

\mypara{Classification results}
\begin{figure*}[ht]
\centering
\subfloat[][]{\includegraphics[scale=0.55]{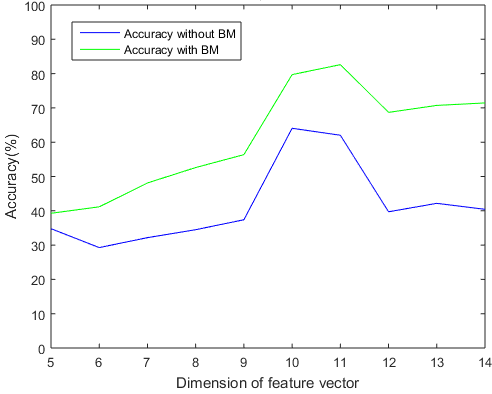}\label{KNN}}\hspace{0.05cm}
\subfloat[][]{\includegraphics[width=.5\linewidth]{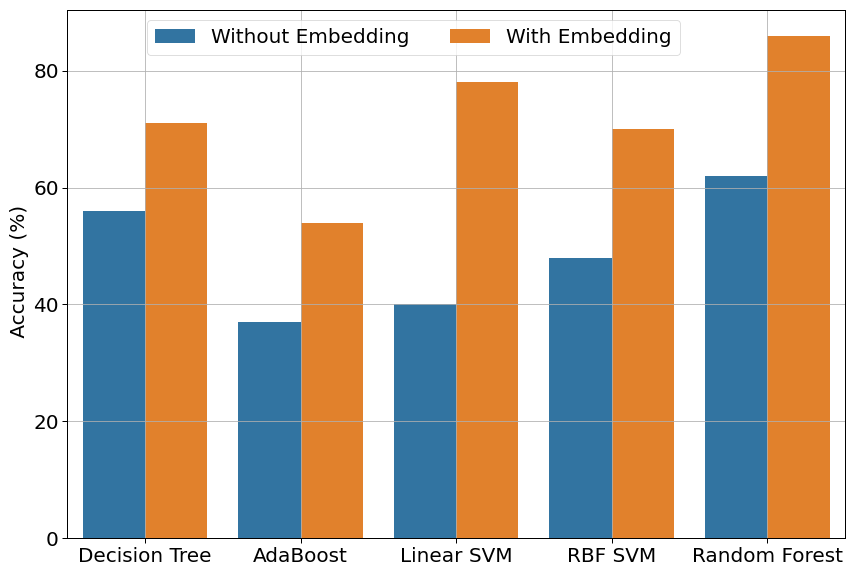}\label{diff}}
        \setlength{\belowcaptionskip}{-10pt}

\caption{Performance of $k$-NN classifier in Euclidean and boosted embedding space as a function of the feature dimension. The second figure shows the performance gain in classification accuracy in embedded space over Euclidean space for different classifiers.}
\label{classification}
\end{figure*}
This section evaluates how effective the boosting-based feature transformation technique is in improving the classification performance of haptic textures. Training signals are labeled by their class, and the training triplets are constructed with two samples from a single class and the third sample from a different one. We seek to show that classes of signals are easily separable in the embedding space and hence perform a better classification task: predicting which of the 69 classes an unlabeled test signal belongs to. From a total of 690 training samples, we generated a total of 6210 triplets and trained metric $M$ with a regularizing parameter $v = 10^{-7}$ in Eq. \ref{primal}, for 3000 iterations.
\footnotetext{Some abbreviation used are as follows: Acc (acceleration), Img (Image), Snd (sound), Frc (friction), CQFB(Constant Q-factor filter bank).}

Our work rests on two key components: (a)~the CQFB(Constant Q-factor bandwidth) features that capture salient features of the acceleration data, and (b)~the boosting-based feature transform that further enhances separability of signals in the projected space. To demonstrate the individual role of each of these factors to the overall efficacy of the method, we conduct various ablation studies in which we choose different features, or omit the metric embedding, or both. The results for the classification of unseen test signals with different features are given in Table~\ref{accuracyTable1}. In the first experimentation, Our boosted features use only a simple $k$-NN ($k = 3$) classifier, to highlight the advantages of the features alone. The classification performance of various other classifiers is presented later in the section.

Table~\ref{accuracyTable1} shows the comparison of $k$-NN classification performance with different features from prior work, in original feature space, and in the learned boosted embedding space. Experimentation is done for various feature dimensions, and the best accuracy is reported for each feature in Table~\ref{accuracyTable1}. As we can see, some of the features such as spectral peak, spectral frequency cannot capture the hidden features of the data well. Note that the CQFB feature with constant bandwidth $(\alpha = 1)$ also fails to capture the complexity of underlying data. This is expected, as the discriminating power is better at a lower frequency, and by using an equal bandwidth filter, we lose signal feature by capturing lower frequency components at a low resolution. Using a variable bandwidth CQFB feature, accuracy improves significantly (from $2.5\%$ to $60\%$). This further validates the significance of using CQFB feature in the classification task. As we can see from Table~\ref{accuracyTable1}, the learned metric consistently improves k-NN classification accuracy for most of the features, and the best accuracy is achieved with the combination of the CQFB feature and boosted embedding.

We also present accuracies with a variety of other features and classification methods from prior work. Table~\ref{accuracyTable2} reports various other classifier's results and its comparison from related work by Strese \etal~\cite{TUM}. In \cite{TUM}, the authors present the classification results of the Naive Bayes classifier and Decision Tree with several features. Note that all unimodal features which are obtained from acceleration data alone have low accuracy ($\leq39\%$), and the accuracy improves to $74\%$ for Naive Bayes classifier and $67\%$ for Decision Tree only when other modalities of data are used for features extraction. In contrast, CQFB feature augmented by boosted embedding technique yields promising results only with acceleration data and is beaten marginally only by the method that resorts to other modalities to improve recognition.

Further, we compare in Fig. \ref{classification} the performance of the learned distance metric with the baseline Euclidean distance metric for classification tasks across various features dimensions using the CQFB feature. It can be observed that the boosted embedding improves the accuracy of the KNN classifier significantly. In Fig. \ref{diff} we study the same for other types of classifiers that one may select.  As we can see in Fig. \ref{diff}, the classification performance of all classifiers improves in the embedding space. This validates that the learned distance metric is much more effective in improving the separability of data than the Euclidean metric.

\begin{figure*}[ht]
  \centering
   \subfloat[][]{\includegraphics[scale=0.35]{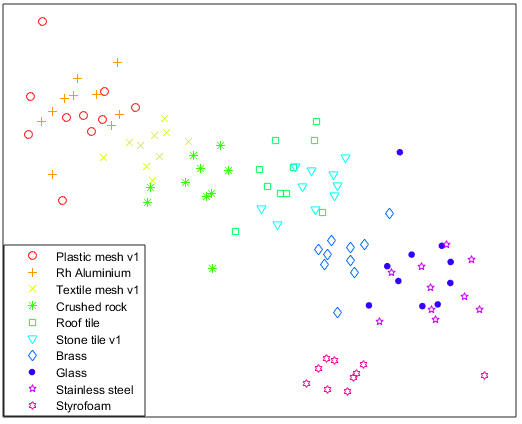}\label{CQFB_alpha1}}\hspace{0.1cm}
  \subfloat[][]{\includegraphics[scale=0.35]{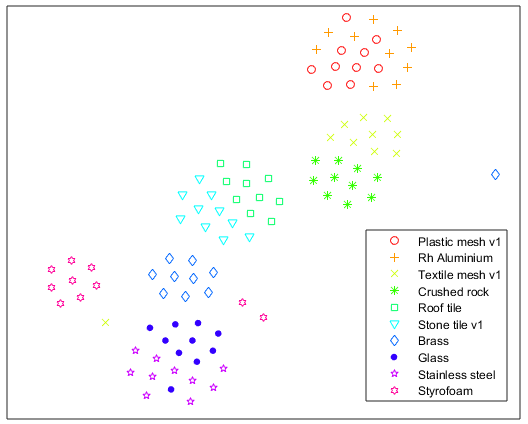}\label{CQFB_alpha1.8}}\hspace{0.1cm}
  \subfloat[][]{\includegraphics[scale=0.35]{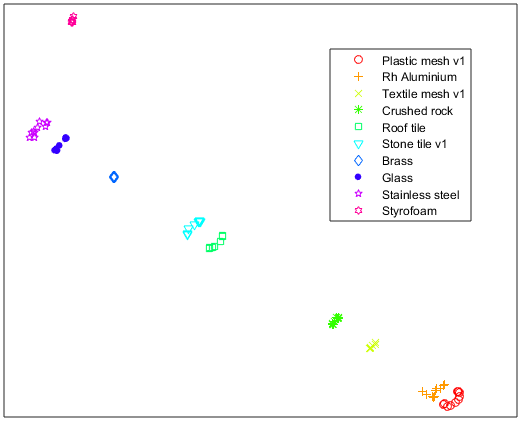}\label{CQFB_MSpace}}\hspace{0.1cm}
          \setlength{\belowcaptionskip}{-10pt}

  \caption{Illustrating improved separability of classes due to the choice of CQFB feature and boosted embedding. Different symbols and colors indicate different materials. Features are projected onto 2D space using t-SNE plot, (a) Constant bandwidth feature (CQFB features with $\alpha = 1$), (b) Variable bandwidth feature (CQFB features with $\alpha = 1.8$) and (c) Variable bandwidth feature (CQFB features with $\alpha = 1.8$) projected after embedding.}
  \label{data_projection}
\end{figure*}

\mypara{Clustering}
We visualize the embedding space (projected to 2D using t-SNE~\cite{tSNE}) in Figure~\ref{data_projection}, with just randomly chosen ten classes out of 69 for clarity. On the left (a), we see constant bandwidth CQFB features ($\alpha = 1$) fail to recover the underlying data structure. In the middle (b), variable bandwidth CQFB features form clear class-based clusters, but they are fairly diffused, and it is easy to see how a simple classifier can confuse similar classes near their class boundaries. On the right (c), boosting-based feature transformation induces more compact clusters with low intra-class variance, but high inter-class separation. Classes are much better separable in Fig. \ref{CQFB_MSpace}.

\mypara{Haptic discrimination} In this section, we evaluate the pairwise distinguishability of surface materials in the embedding space. We measure the pairwise distinguishability of two surface materials as the average number of triplet $(x_{i}, x_{j}, x_{k})$ comparisons for which the distance between intra-class samples $x_i$ and $x_j$ is lesser than the distance between inter-class samples $x_i$ and $x_k$ $(d_{M}(x_{i}, x_{j}) < d_{M}(x_{i}, x_{k})$. We define the dissimilarity index $p_{ab}$ for each material pair $(a, b)$ as follows:

\begin{equation} \label{discrimination}
p_{bc} = \frac{|S|}{|C_{bc}|};
S = \{(x_{i}, x_{j}, x_{k}) | d_{M}(x_{i}, x_{j}) < d_{M}(x_{i}, x_{k})\}
\end{equation}

\begin{figure*}[!t]
\centering
\subfloat[Euclidean]{\includegraphics[scale=0.35]{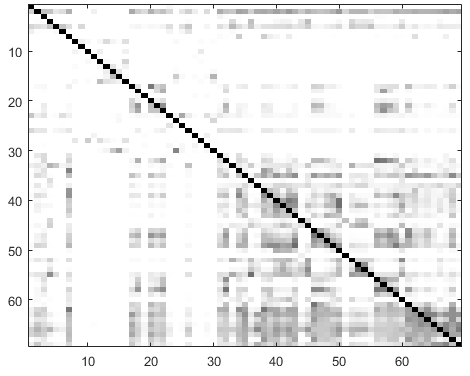}\label{fig:euclidean}}\hspace{5mm}
\subfloat[Ours]{\includegraphics[scale=0.35]{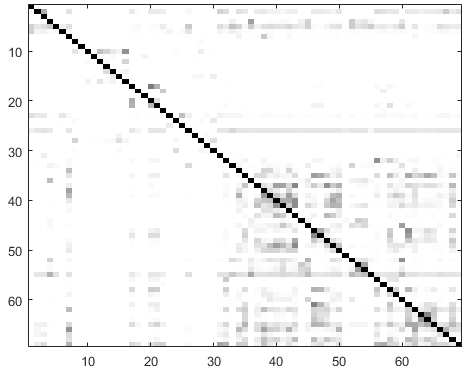}\label{fig:boost}}\hspace{5mm}
\subfloat[Perceptual]{\includegraphics[scale=0.35]{result/BoostMetric.png}\label{fig:perceptual}}
          \setlength{\belowcaptionskip}{-10pt}

\caption{Confusion matrix ($69 \times 69$ matrix) while discriminating different texture pairs obtained with (a) Euclidean metric, (b) Boosting-based Mahalanobis metric, c) Ground-truth perceptual distance obtained from human perceptual judgments~\cite{TUM}. Darker spots indicate more confusion in distinguishing a pair of textures. The matrix shows many indistinguishable textures pairs in the Euclidean space.  Boosting improves the separability of surface textures. Many texture pairs (such as brick (index = 8) and granite (index = 10)) that are difficult to discriminate for users are readily distinguishable by the learned metric, while some texture pairs (plastic mesh (index = 4) and carpet (index = 36) ) that are easy to discriminate by humans were more challenging for the learned metric.}
\label{fig:pairwise_distinguish}
\end{figure*}

where $(x_{i}, x_{j}, x_{k})$ is a triplet constructed from two signals $x_{i}$ and $x_{j}$ of same class $b$, and $x_{k}$ from different class $c$ and $C_{bc}$ is a set of all possible triplets constructed from signals of class $b$ and $c$ without any constraint on the distance. The dissimilarity index ($p_{bc}$) is a measure of the probability of distinguishing two materials in the embedded space using the learned distance matrix $M$. A small value of the dissimilarity index indicates a more confusing material pair. Figure \ref{fig:euclidean} and \ref{fig:boost} shows the comparison between confusion matrices obtained using Euclidean and boosting with CQFB features. The darker spots in the confusion matrix indicate more confusion in distinguishing texture pairs. As we can see, materials are better separable in the learned embedding space than in the Euclidean space. The overall discrimination error is obtained by finding the mean of the confusion index of all material pairs. The average discrimination error in separating a  material-pair using Euclidean metric and Mahalanobis metric are $7.10\%$ and $3.76\%$, respectively, substantiating the usefulness of boosted embedding in improving separability of two classes of materials.

\mypara{Limitation} We evaluate our model by comparing its performance with actual human subject performance in discriminating haptic textures. For human perceptual similarity feedback, we have used the similarity matrix $\mathbf{S_{H}}$ provided by authors in \cite{TUM}. Each element $\mathbf{S_{H}}(i, j)$ denotes the number of subjects (out of 30) having perceived different materials $i$ and $j$ to be similar and clubbed them into the same group. In other words, each element in the matrix $\mathbf{S_{H}}$ indicates the confusion in separating two materials apart. As we can see in Figure \ref{fig:boost} and \ref{fig:perceptual}, the dissimilarity trend in both semantic and perceptual distance matrices is quite different, indicating the semantic metric’s inability to represent perceptual dissimilarity. We believe this is due to not incorporating human perceptual limitation in distinguishing two very similar signals (called just noticeable difference-JND) in the modeling process. The inclusion of human perceptual limitations in terms of JND in the distance metric learning process will definitely increase its usability.

\tightsection{Conclusions}\label{sec:conclusion}

In this work, we have presented a framework for the discriminative analysis of haptic textures using a learned distance metric. We demonstrated the effectiveness of hand-crafted spectral features in improving the distinguishability of real-world surface textures. Further, we observed that the learned boosted embedding on the CQFB space using class-based supervision improves separability between haptic signals. We found that our method outperforms the state-of-the-art method by a large margin in classification tasks when only the single-exploration acceleration data is used. The extensive experimentation with different predictors and feature dimensions demonstrates the effectiveness of our approach for several downstream tasks such as texture recognition and discrimination. Our current work mainly looks into class-based separability between objects. It will be intriguing to study how effectively the boosting-based technique can model human perceptual limitations such as JND.

% \section*{ACKNOWLEDGMENT}

% 

\let\oldbibliography\thebibliography
\renewcommand{\thebibliography}[1]{%
  \oldbibliography{#1}%
  \setlength{\itemsep}{0pt}%
}
%   \bibliographystyle{abbrv}
% \bibliography{references} 
% \bibliographystyle{IEEEtran}

% \vspace{-5px}
\bibliographystyle{savetrees}

\bibliography{reference}

\begin{thebibliography}{10}
\providecommand*{\selectlanguage}[1]{\relax}%
\providecommand*{\savetreesbibnote}[1]{#1}%

\bibitem{psychophysics1}
S.~Bensma{\"\i}a and M.~Hollins.
\newblock Pacinian representations of fine surface texture.
\newblock \emph{Perception \& psychophysics}, 2005.

\bibitem{freqFeature}
K.~Beyer, J.~Goldstein, et~al.
\newblock When is “nearest neighbor” meaningful?
\newblock In \emph{ICDT}. 1999.

\bibitem{Chathuranga}
D.~S. Chathuranga, Z.~Wang, et~al.
\newblock A biomimetic soft fingertip applicable to haptic feedback systems for
  texture identification.
\newblock In \emph{HAVE}. 2013.

\bibitem{Culbertson}
H.~Culbertson, J.~Unwin, et~al.
\newblock Modeling and rendering realistic textures from unconstrained
  tool-surface interactions.
\newblock \emph{ToH}, 2014.

\bibitem{colGen}
A.~Demiriz, K.~P. Bennett, et~al.
\newblock Linear programming boosting via column generation.
\newblock \emph{Machine Learning}, 2002.

\bibitem{phonemes}
M.~Enriquez, K.~MacLean, et~al.
\newblock Haptic phonemes: basic building blocks of haptic communication.
\newblock In \emph{ICMI}. 2006.

\bibitem{Fishel}
J.~A. Fishel and G.~E. Loeb.
\newblock Bayesian exploration for intelligent identification of textures.
\newblock \emph{Frontiers in neurorobotics}, 2012.

\bibitem{gao2016deep}
Y.~Gao, L.~A. Hendricks, et~al.
\newblock Deep learning for tactile understanding from visual and haptic data.
\newblock In \emph{ICRA}. 2016.

\bibitem{NCA}
J.~Goldberger, G.~E. Hinton, et~al.
\newblock Neighbourhood components analysis.
\newblock In \emph{NeurIPS}. 2005.

\bibitem{Jamali}
N.~Jamali and C.~Sammut.
\newblock Majority voting: Material classification by tactile sensing using
  surface texture.
\newblock \emph{ToR}, 2011.

\bibitem{DFT321}
N.~Landin, J.~M. Romano, et~al.
\newblock Dimensional reduction of high-frequency accelerations for haptic
  rendering.
\newblock In \emph{EuroHaptics}. 2010.

\bibitem{lederman90haptic}
S.~J. Lederman and R.~L. Klatzky.
\newblock Haptic classification of common objects: Knowledge-driven
  exploration.
\newblock \emph{Cognitive psychology}, 1990.

\bibitem{tSNE}
L.~v.~d. Maaten and G.~Hinton.
\newblock Visualizing data using t-sne.
\newblock \emph{JMLR}, 2008.

\bibitem{priyadarshini2019perceptnet}
K.~Priyadarshini, S.~Chaudhuri, et~al.
\newblock Perceptnet: Learning perceptual similarity of haptic textures in
  presence of unorderable triplets.
\newblock In \emph{WHC}. 2019.

\bibitem{richardson2019improving}
B.~A. Richardson and K.~J. Kuchenbecker.
\newblock Improving haptic adjective recognition with unsupervised feature
  learning.
\newblock In \emph{ICRA}. 2019.

\bibitem{Katherine14}
J.~M. Romano and K.~J. Kuchenbecker.
\newblock Methods for robotic tool-mediated haptic surface recognition.
\newblock In \emph{Haptics Symposium}. 2014.

\bibitem{boost}
C.~Shen, J.~Kim, et~al.
\newblock Positive semidefinite metric learning with boosting.
\newblock In \emph{NeurIPS}. 2009.

\bibitem{Sinapov}
J.~Sinapov, V.~Sukhoy, et~al.
\newblock Vibrotactile recognition and categorization of surfaces by a humanoid
  robot.
\newblock \emph{ToR}, 2011.

\bibitem{optimal_perception_freq}
L.~Skedung, M.~Arvidsson, et~al.
\newblock Feeling small: exploring the tactile perception limits.
\newblock \emph{Scientific reports,Nature}, 2013.

\bibitem{strese2019haptic}
M.~Strese, L.~Brudermueller, et~al.
\newblock Haptic material analysis and classification inspired by human
  exploratory procedures.
\newblock \emph{ToH}, 2019.

\bibitem{WHC15}
M.~Strese, C.~Schuwerk, et~al.
\newblock Surface classification using acceleration signals recorded during
  human freehand movement.
\newblock In \emph{WHC}. 2015.

\bibitem{TUM}
---.
\newblock Multimodal feature-based surface material classification.
\newblock \emph{ToH}, 2017.

\bibitem{LMCA}
L.~Torresani and K.-c. Lee.
\newblock Large margin component analysis.
\newblock In \emph{NeurIPS}. 2007.

\bibitem{andrew}
E.~P. Xing, M.~I. Jordan, et~al.
\newblock Distance metric learning with application to clustering with
  side-information.
\newblock In \emph{NeurIPS}. 2003.

\bibitem{Zheng}
H.~Zheng, L.~Fang, et~al.
\newblock Deep learning for surface material classification using haptic and
  visual information.
\newblock \emph{ToM}, 2016.

\end{thebibliography}

\end{document}